\documentclass[conference]{IEEEtran}
\IEEEoverridecommandlockouts
\usepackage[numbers]{natbib}
\usepackage{amsmath,amssymb,amsfonts}
\usepackage{graphicx}
\usepackage{textcomp}
\usepackage{xcolor}
\usepackage{balance} % for balancing columns on the final page

\usepackage{graphicx}
\usepackage{booktabs}
\usepackage{float}
\usepackage{hyperref}
\usepackage{algorithm}
\usepackage{algorithmicx,algpseudocode}
\usepackage{multirow}
\usepackage{algpseudocode}
\usepackage{booktabs}

\def\NoNumber#1{{\def\alglinenumber##1{}\State #1}\addtocounter{ALG@line}{-1}}
\usepackage{multirow,color,amsmath}
\usepackage{array} % DO NOT CHANGE THIS
\usepackage{tikz,balance}
\usepackage{enumitem,xcolor,booktabs,colortbl}
\newcommand\rew[1]{\textcolor{red}{#1}}

\newcommand{\sS}{\mathcal{S}}
\newcommand{\sA}{\mathcal{A}}
\newcommand{\sP}{\mathcal{P}}
\newcommand{\sR}{\mathcal{R}}
\newcommand{\sC}{\mathcal{C}}
\def\BibTeX{{\rm B\kern-.05em{\sc i\kern-.025em b}\kern-.08em
    T\kern-.1667em\lower.7ex\hbox{E}\kern-.125emX}}

\DeclareRobustCommand*{\IEEEauthorrefmark}[1]{%
    \raisebox{0pt}[0pt][0pt]{\textsuperscript{\footnotesize\ensuremath{#1}}}}
\begin{document}

\title{Evolving Constrained \\Reinforcement Learning Policy}
% \author{\IEEEauthorblockN{Anonymous Authors}}

\author{\IEEEauthorblockN{Chengpeng Hu\IEEEauthorrefmark{1,2}, Jiyuan Pei\IEEEauthorrefmark{1,2}, Jialin Liu\IEEEauthorrefmark{2,1}, and Xin Yao\IEEEauthorrefmark{2,1}}

\IEEEauthorblockA{\IEEEauthorrefmark{1}Research Institute of Trustworthy Autonomous Systems (RITAS),\\ Southern University of Science and Technology, Shenzhen, China.\\
}
\IEEEauthorblockA{\IEEEauthorrefmark{2}Guangdong Key Laboratory of Brain-inspired Intelligent Computation, 
Department of Computer Science and Engineering, \\Southern University of Science and Technology, Shenzhen, China.\\
hucp2021@mail.sustech.edu.cn, peijy2020@mail.sustech.edu.cn, liujl@sustech.edu.cn, xiny@sustech.edu.cn}

\thanks{This work is accepted by the 2023 International Joint Conference on Neural Networks (IJCNN).}
}
% \author{\IEEEauthorblockN{1\textsuperscript{st} Chengpeng Hu}
% \IEEEauthorblockA{\textit{Dept. of Computer Science and Engineering} \\
% \textit{Southern University of Science and Technology }\\
% shenzhen, China \\
% hucp2021@mail.sustech.edu.cn}
% \and
% \IEEEauthorblockN{2\textsuperscript{nd} Jiyuan Pei}
% \IEEEauthorblockA{\textit{Dept. of Computer Science and Engineering} \\
% \textit{Southern University of Science and Technology }\\
% shenzhen, China \\
% peijy2020@mail.sustech.edu.cn}
% \and
% \IEEEauthorblockN{3\textsuperscript{rd} Jialin Liu}
% \IEEEauthorblockA{\textit{Dept. of Computer Science and Engineering} \\
% \textit{Southern University of Science and Technology }\\
% shenzhen, China \\
% liujl@sustech.edu.cn}
% \and
% \IEEEauthorblockN{4\textsuperscript{th} Xin Yao}
% \IEEEauthorblockA{\textit{Dept. of Computer Science and Engineering} \\
% \textit{Southern University of Science and Technology }\\
% shenzhen, China \\
% xiny@sustech.edu.cn}
% \thanks{\rew{This work was supported by the Research Institute of Trustworthy Autonomous Systems (RITAS), the Guangdong Provincial Key Laboratory (Grant No. 2020B121201001), the Program for Guangdong Introducing Innovative and Enterpreneurial Teams (Grant No. 2017ZT07X386), the Shenzhen Science and Technology Program (Grant No. KQTD2016112514355531) and the National Natural Science Foundation of China (Grant No. 61906083).}}
% }

\maketitle

\begin{abstract}
Evolutionary algorithms have been used to evolve a population of actors to generate diverse experiences for training reinforcement learning agents, which helps to tackle the temporal credit assignment problem and improves the exploration efficiency. However, when adapting this approach to address constrained problems, balancing the trade-off between the reward and constraint violation is hard. In this paper, we propose a novel evolutionary constrained reinforcement learning (ECRL) algorithm, which adaptively balances the reward and constraint violation with stochastic ranking, and at the same time, restricts the policy's behaviour by maintaining a set of Lagrange relaxation coefficients with a constraint buffer. Extensive experiments on robotic control benchmarks show that our ECRL achieves outstanding performance compared to state-of-the-art algorithms. Ablation analysis shows the benefits of introducing stochastic ranking and constraint buffer. 
% Codes are available at \url{https://anonymous.4open.science/r/Evolutionary-Constrained-Reinforcement-Learning-25E1/}
\end{abstract}


%%%%%%%%%%%%%%%%%%%%%%%%%%%%%%%%%%%%%%%%%%%%%%%%%%%%%%%%%%%%%%%%%%%%%%%%
\section{Introduction}
\label{sec:intro}

Reinforcement learning (RL) has shown strong competence in several domains, such as games~\cite{mnih2015human,silver2016mastering,hu2022dorl}. However, RL algorithms suffer from the temporal credit assignment problem with sparse reward, instability and inefficient sampling that often occur in real-world problems~\cite{sutton2018reinforcement,khadka2018evolution}. To tackle these issues, \citet{khadka2018evolution} proposed a hybrid algorithm, called evolutionary reinforcement learning (ERL). ERL evolves a population of actors which are used to generate diverse experiences for training an RL agent and periodically injects the agent's gradient information into the corresponding evolutionary algorithm (EA). ERL outperforms some state-of-the-art RL algorithms and EA on some unconstrained robot control problems~\cite{khadka2018evolution}.

Most RL algorithms are based on the Markov decision process (MDP)~\cite{sutton2018reinforcement}, which does not consider any constraint at all. When addressing real-world problems with constraints, such as constrained robot control problems~\cite{tessler2018reward} and autonomous vehicle~\cite{shalev2016safe},
% and logistics~\cite{delarue2020reinforcement}, 
an RL agent may learn a policy that achieves a high reward but violates one or more constraints.

Though ERL was originally not designed for constrained optimisation problems, the experience diversity introduced by ERL has the potential to achieve higher performance in handling constrained problems. It is intuitive to combine ERL and constraint reinforcement learning (CRL) approaches to address constrained problems. However, when adopting both classic and state-of-the-art CRL approaches, such as the reward shape method~\cite{ng1999policy} and reward constrained policy optimisation (RCPO)~\cite{tessler2018reward}, to ERL, the dilemma between reward and constraint violation is observed (cf. Fig. \ref{fig:reshape}). This can be explained by the conflicting behaviour of EA and RL. 
As the CRL approach manipulates the RL agent only, the RL agent learns the policy with both reward and constraints while the EA evaluates and selects actors considering reward only~\cite{khadka2018evolution}. Thus, the agent, expected to maximise the reward and minimise the constraint violation, learns from experiences sampled by the actors of the EA's population, which ignore constraints. More specifically, the EA and RL components of ERL sample experiences from two distributions with different concerns, one over reward only and the other over both reward and constraints. When most of the experience buffer is reward-only experiences, the gradient will be directed to optimise reward only. 
The dilemma between the reward and constraint violation as well as the presence of the conflicting behaviour of EA and RL suggests that simply applying CRL techniques to ERL is not enough. The actor selection in EA should consider both rewards and constraints. 
%experimently demonstrate the failure of ERL using some classic and state-of-the-art CRL approaches on robot control benchmarks to support our motivation.

%Additional manipulation is needed to tackle the conflicting behaviours of EA and learner, and balance the trade-off between reward and constraint violation.

%This motivates us to propose ECRL which leverages stochastic ranking and Lagrange relaxation.

% However, when applying those methods to ERL, the trade-off between reward and constraint violation is observed. As shown in Fig. \ref{fig:reshape}, along with the augmentation of raw reward value obtained by the learnt policy, the corresponding constraint violation fluctuates. In ERL, the actors are selected by an EA based on the reward only while the learner considers both reward and constraint violations. 

% a conflict raised when we apply these methods to handle constraints with the introduce of Evolutionary Reinforcement Learning (ERL).  Evolutionary Reinforcement learning (ERL)~\cite{khadka2018evolution} provides  and injects gradient information into by an RL learning. Higher sample efficiency and faster learning are presented by this framework compatible with off-policy RL algorithms. It is reported that ERL can help to overcome temporal credit assignment with sparse reward, instability and inefficient sampling that troubles vanilla RL algorithms.
%Since the objective of learning in constrained problems considers both reward and constraints, 

%We propose to address this through stochastic ranking~\cite{runarsson2000stochastic}.
In this paper, we propose to consider both reward and constraints during actor selection through stochastic ranking~\cite{runarsson2000stochastic} and design an evolutionary constrained reinforcement learning (ECRL) algorithm. ECRL inherits remarkable properties of ERL~\cite{khadka2018evolution} and leverages stochastic ranking~\cite{runarsson2000stochastic} to rank the actors while balancing their rewards and constraint violations. Additionally, to provide diverse experiences with constraint information for the RL agent, ECRL adopts the Lagrangian relaxation method and introduces a multiplier for each individual in the population. The multipliers are updated and improved using an extra constraint buffer that stores historical episodic constraints. 
%Notably, the multiplier works on providing diverse experiences with constraint information for RL learner, which does not affect stochastic ranking. 

The contributions of this paper are summarised as follows:
% (i) We introduce stochastic ranking and Lagrange multiplier to ERL, which enables ERL to deal with constrained problems; (ii) A constraint buffer is also introduced, which maintains a set of historically good Lagrangian multipliers that the RL agent can consult every now and then when selecting actors. 
% (iii) Extensive experiments over robotic control benchmarks are conducted and demonstrate that our ECRL performs better than several state-of-the-art algorithms including ERL, variants of ERL using different CRL approaches, and a state-of-the-art algorithm for constrained problems. An ablation study is also presented. ECRL requires no prior knowledge and is easy to implement.
\begin{itemize}
    \item We determine the phenomenon of the conflicting behaviour that occurs when incorporating constrained RL into ERL. The necessary coordination of evolutionary constraint handling techniques is emphasised.
    \item  We introduce stochastic ranking and Lagrange multiplier to ERL, which enable ERL to deal with constrained problems and tackle the conflicting behaviour.
    \item  A constraint buffer is introduced, which maintains a set of historically Lagrangian multipliers that the RL agent can consult every now and then when selecting actors.
    \item Experiments conducted over robotic control benchmarks validate the occurrence of conflicting behaviours and demonstrates that our ECRL performs better than several state-of-the-art algorithms including ERL, variants of ERL using different CRL approaches, and state-of-the-art algorithm for constrained problems. An ablation study is also presented. ECRL requires no prior knowledge and is easy to implement.
\end{itemize}

%propose a novel evolutionary constrained reinforcement learning (ECRL) algorithm. ECRL models each individual as an actor. Behaviours of agents can be restricted with Lagrange relaxation multipliers. Stochastic ranking is adopted to rank the individuals while balancing the reward and constraints.
%(ii) A constraint buffer is introduced to store historical constraint values for the stable update and improvement of multipliers. 

\def\nouse{
\section{Motivation}
Some questions and answers for the proposed method. This part may not appear in the final article but still be meaningful for introspection.
\begin{itemize}
    \item Why do we want to apply ERL to this kind of task?
    \begin{itemize}
         \item ERL is easy to implement. Efficient sampling and faster learning. ERL can be applied to almost all off-policy RLs to improve the performance of base algorithm.
    \item ERL inherits the advantages of EA. These advantage brought by EA can help to overcome:
    \begin{itemize}
        \item Temporal credit assignment with sparse rewards
        \item Instability
        \item Inefficient sampling (exploration)
    \end{itemize}
    \item ERL may help to solve Real-world problems which have sparse reward and many constraints. (sparse reward maybe not discussed in this paper)
    
    \end{itemize}
   \item However, ERL can not be applied to this task directly because it does not consider constraints at all. If we want to use the Pros of ERL, we have to find some way to make ERL to adapt to this task.
   \item To the best of our knowledge, ECRL has been not addressed in the literature.
   \item Why can exist methods not directly helps?
   \begin{itemize}
       \item Conflict of EA and RL
       \item Current methods only improve RL part. In ERL, most experiences that learner learns from are sampled by EA part, which using reward as fitenss for selection. Even SOTAs restrict the learner's policy, it still fail without constraint mechanism for EA.
       \item So we use stochastic ranking for balance. The constrained policy with better reward holds the larger chance to be selected.
   \end{itemize}
    \item Why Stochastic ranking?
    \begin{itemize}
        \item Similar answer with above questions
        \item list some Pros of stochastic ranking
    \end{itemize}
    \item Why evolving $\lambda$ even if we have stochastic ranking?
    \begin{itemize}
        \item Stochastic mainly helps EA. The experience collected will affect RL indirectly.
        \item Evolving $\lambda$ works for RL.
        \begin{itemize}
            \item Faster learning
            \item Similarly to multi timescale approach
                \begin{itemize}
                    \item Difference? We reserve previous multipliers
                    \item Convergence proof? Not sure how to start, may follow timescale approach
                \end{itemize}
        \end{itemize}
        \item These two parts complement each other (coordination)
        \item Ablation experiments are needed
    \end{itemize}
\end{itemize}}

\section{Background}\label{sec:rw}

\subsection{Constrained Reinforcement Learning}
% When addressing real-world problems with constraints, such as autonomous vehicle~\cite{shalev2016safe} and logistics~\cite{delarue2020reinforcement}, an RL agent may learn a policy that achieves a high reward but violates one or more constraints~\cite{achiam2017constrained}.
%\subsection{Constrained Markov Decision Process}
%Extending the Markov decision process~\cite{sutton2018reinforcement}, 
%We adapt the notations used in \cite{liu2021policy}.
To address constrained optimisation problems, the MDP~\cite{sutton2018reinforcement} is extended to the constrained MDP (CMDP)~\cite{altman1999constrained}. On the basis of the maximising discounted cumulative reward, CMDP restricts policy under constraints. 
%A classic approach to address CMDP is the reward shaping method~\cite{ng1999policy}, which shapes the reward with a weighted penalty terms to penalise constraint violations and guide the learning. Recently, more sophisticated approaches were proposed under different assumptions or characteristics of the constrained optimisation problems~\cite{liu2021policy,achiam2017constrained}, such as using Lagrangian relaxation~\cite{altman1999constrained,chow2017risk,tessler2018reward,stooke2020responsive}, augmenting the objective function by logarithmic barrier functions~\cite{liu2020ipo}, solving the quadratic constrained problem approximated from the original problem~\cite{achiam2017constrained}, and projecting an infeasible policy to its closest feasible policy~\cite{Yang2020Projection-Based}. 

The CMDP~\cite{altman1999constrained,liu2021policy} is defined as a tuple $(\sS,\sA,\sR,\sC,\sP,\gamma)$, where $\sS$ is the set of states, $\sA$ is the set of actions, $\sR:\sS\times \sA \times \sS \mapsto \mathbb{R}$ is the reward function. $\sC$ is the cost function of a constraint with $\sC:\sS\times \sA \times \sS \mapsto \mathbb{R}$. $\sP:\sS\times \sA \times \sS \mapsto [0,1]$ is the transition probability function. 
%$\mu:\sS\mapsto[0,1]$ is the starting state distribution and 
$\gamma$ is the discount factor. 
A policy $\pi:\sS\mapsto \sP(\sA)$ is a mapping from states to a probability distribution over actions. $\pi(a_t|s_t)$ is the probability of taking action $a_t$ in state $s_t$ at time $t$. 
%Usually, a constraint $\sC_i = g(c(s_t,a_t),c(s_{t+1},a_{t+1}),\dots)$ is restricted by threshold $\alpha$. 
Usually, a cumulative constraint $\sC = g(c(s_0,a_0,s_{1}),...,c(s_{t},a_{t},s_{t+1}))$ is restricted by a threshold $\epsilon$ where $c(s,a,s')$ is a per-step penalty to constraint violation.
$J^{\pi}_{\sC}$ denotes the expectation of the cumulative constraint and is formulated as $J^{\pi}_{\sC} = \mathbb{E}_{\tau \sim \pi}[\sC]$, where $\tau$ denotes a trajectory $(s_0,a_0,s_1,a_1,\dots)$ and $\tau \sim\pi$ denotes trajectories sampled from $\pi$.
%\todo[inline]{The above formulation seems to be wrong/imcomplete}
The goal of CMDP is to find a policy $\pi_\theta$ that maximises the discounted cumulative reward subjecting to some constraints, formulated as~\cite{altman1999constrained}:
\begin{eqnarray}
\max_{\theta}~J_{\sR}^{\pi_\theta}&=&\mathbb{E}_{\tau \sim\pi_\theta}[\sum_{t=0}^{\infty} \gamma^t \sR(s_t,a_t,s_{t+1})]  \\
s.t.&& J^{\pi}_\sC\leq \epsilon.
\end{eqnarray}

Several constrained reinforcement learning (CRL)~\cite{altman1999constrained} approaches have been proposed to handle constraints while maximising cumulative discounted rewards~\cite{garcia2015comprehensive,liu2021policy}.
%, as reviewed in the work of~\cite{garcia2015comprehensive,liu2021policy}. 
%Latest advances in CRL have been reviewed in \cite{garcia2015comprehensive,liu2021policy}. 
Augmenting the objective function in case of constraint violation is a commonly used category~\cite{borkar2005actor,tessler2018reward,calian2021balancing}.
%category of commonly used CRL approaches~\cite{borkar2005actor,tessler2018reward,calian2021balancing}. 
The reward shape~\cite{ng1999policy} is a classic and easy-to-implement method for constrained optimisation. It shapes the reward function with a weighted penalty term.
%with a constant constraint coefficient.
An actor-critic algorithm using a Lagrange multiplier proposed by~\citet{borkar2005actor} can be regarded as a primal-dual type learning algorithm using different timescales. \citet{tessler2018reward} proposed a multi-timescale approach, reward constrained policy optimisation (RCPO). Instead of using a constant value, the penalty coefficient is updated during learning in a slower time scale than policy gradient update~\cite{tessler2018reward}. \citet{calian2021balancing} extended the Lagrange formulation to nested optimisation problems via meta-gradients. PID Lagrangian, presented by \citet{stooke2020responsive}, considers Lagrangian multiplier update as a dynamic control problem. This method~\cite{stooke2020responsive} avoids oscillating cases when updating the multiplier with extra parameters. The interior-point policy optimisation (IPO)~\cite{liu2020ipo} augments the objective function by logarithmic barrier functions. It is assumed that policies must be feasible during initialisation~\cite{liu2021policy}.
%with logarithmic barrier functions with performance guarantee bound given, which is easy to implement and be tuned. 
%The aforementioned approaches mainly deal with constraint by augmenting objective function, which is common and popular in this area. However, there is another way called as trust region type approach~\cite{schulman2015trust}.

Another popular category of CRL approaches is the trust region method.
Constrained policy optimisation (CPO)~\cite{achiam2017constrained} searches constrained policy locally. The best-improved policy with constraint satisfaction is selected after updating with the approximated prediction of constraint violations. Different to CPO, projection-based constrained policy (PCPO)~\cite{Yang2020Projection-Based} uses two steps to optimise the constrained policy iteratively. PCPO first learns a reward-preferred policy locally, and then projects this policy back into the constrained region~\cite{Yang2020Projection-Based}. 

\rew{
% Although aforementioned approaches may work well with vanilla RL algorithm, they hardly directly cooperate with ERL. It is expected to leverage remarkable abilities of ERL 
% into constraint RL problems.
% Some of the aforementioned approaches rely on certain assumptions of the problem to be solved~\cite{liu2021policy} or suffer from sample and calculation complexity~\cite{Yang2020Projection-Based}. To the best of our knowledge, no existing work has applied ERL to address CRL problems, inheriting its remarkable abilities.
% In this work, we incorporate RCPO~\cite{tessler2018reward}, a Lagrange relaxation type method, into ERL since it is easy-to-implement and makes no assumption on the problems.
}
Some of the aforementioned approaches rely on certain assumptions of the problem to be solved~\cite{liu2021policy} or suffer from sample and calculation complexity~\cite{Yang2020Projection-Based}, while RCPO is easy-to-implement and makes no assumption about the problems.
 % In this work, we incorporate RCPO~\cite{tessler2018reward}, a Lagrange relaxation type method into ERL. 

%Among the above approaches, trust region methods IPO is under some certain assumptions~\cite{liu2021policy}. CPO and PCPO suffer from sample and calculation complexity~\cite{Yang2020Projection-Based}. So we incorporate RCPO, a Lagrange relaxation type method into ECRL since it is easy-to-implement and convenient to adapt.

% \rew{They usually face the challenge of sample and calculation complexity~\cite{Yang2020Projection-Based}.}

\subsection{Leverage Reinforcement Learning with Evolution}
EAs have been successfully applied to solve various RL problems. A comprehensive survey is referred to ~\cite{sigaud2022combining}.

Evolutionary strategies (ES)~\cite{rechenberg1965cybernetic}, a type of EAs for numerical optimisation, are often used to train neural network policies for RL tasks such as Atari games~\cite{salimans2017evolution,yang2022evolutionary} and unit commitment problem \cite{liu2014meta}. Instead of calculating the gradients, ES directly searches for the optimal weights of neural networks, where each individual represents the parameters of a network. The best individual in the evolved population is often selected as the final agent.

\citet{khadka2018evolution} also used an EA to evolve weights of neural networks, however, different to the aforementioned works, none of those networks is directly used as the final agent. In other words, the EA is not used to directly search for optimal policies. ERL evolves a population of parameterised actors instead of whole RL agents~\cite{khadka2018evolution}. At each generation, the actors interact with the environment to collect experiences. Actors with better fitness values, calculated based on rewards, have a larger probability to be selected as parents. Mutation and crossover operators are applied to the selected individuals to reproduce new individuals (i.e., actors). An additional RL agent, called ``learner'' in the work of \cite{khadka2018evolution}, is trained simultaneously using the experiences sampled by the actors in the population for better diversity. Periodically, the learner injects its gradient information into the EA by replacing the worst individual with its actor for better convergence~\cite{khadka2018evolution}. The remarkable performance of ERL was observed compared with other RL algorithms, as well as its efficient computation allocation and parallel learning~\cite{khadka2018evolution,khadka2019collaborative}. 

CEM-RL~\cite{pourchot2018cemrl} combines cross-entropy method (CEM) with delayed deep deterministic policy gradient algorithm (TD3). Instead of using genetic operators, CEM-RL samples individuals from the estimated distribution and no actor executes during the process. \citet{bodnar2020proximal} improve the genetic operators with local replay memories and critic of RL. Genetic algorithm (GA) is also combined with an RL algorithm~\cite{marchesini2020genetic} where individuals are only generated by noisy mutation.

ERL \cite{khadka2018evolution} tackles the temporal credit assignment problem and improves exploration efficiency. 
%It is expected to leverage remarkable abilities of ERL  to solve the problems.
It is also worth mentioning that the work of~\cite{hu2022constrained} presents a differential evolutionary algorithm with Q-learning, which uses Q-learning to choose the genetic operator during optimisation. The work of \cite{hu2022constrained} considers evolutionary continuous constraint optimisation while our approach focuses on CRL, which is a sequential decision problem. Although sharing a similar title, our work and \cite{hu2022constrained} actually differ in the problem setting and algorithm. To our best knowledge, none of the aforementioned work based on ERL addresses CRL problems considered in this paper. 
% which is different from the algorithm and problem although it shares a similar title with ours. 
% \todo[inline]{How different? be precise. By ``differential evolutionary algorithm with Q-learning for evolutionary constrained optimisation'' someone may still consider this work similar to the ours. also why didn't we compare to this one? There would be many doubts.}

%\subsection{Evolutionary Reinforcement Learning}
% As presented previously, the evolutionary reinforcement learning (ERL)~\cite{khadka2018evolution} was shown to outperform some state-of-the-art algorithms on some robot control benchmarks. Its mechanism is detailed in \cite{khadka2018evolution} and summarised as follows. The EA's population consists of parameterised actors. At each generation, the actors interact with the environment to collect experience. Actors with better fitness based on rewards have larger probability to be selected as parents. Mutation and crossover operators are applied to selective offsprings to vary parameters of policies. An independent RL agent, called \emph{learner}, learns to act from experiences sampled by the actors maintained by the EA. Periodically, the learner substitutes the individual with the worst fitness in the population to inject its gradient information into the EA for better convergence. 
% \todo[inline]{merge this pari to related work?}

%We focus on the later category.

% \todo[inline]{Trust region}

%To the best of our knowledge, our ECRL for  solving RL problems under constraints has not been addressed in the literature.
\subsection{Stochastic Ranking}
Stochastic ranking~\cite{runarsson2000stochastic} has been shown to be effective in evolutionary constrained optimisation~\cite{runarsson2003constrained}, combinatorial optimisation~\cite{tang2009memetic} and multi-objective optimisation~\cite{li2016stochastic}. It ranks feasible and infeasible individuals in the population according to both their fitness values and penalties. 
%Algorithm~\ref{alg:sr} presents the pseudo code of stochastic ranking. 
%Assuming a solution is a policy in RL, $f(\pi)$ in Algorithm~\ref{alg:sr} denotes the fitness function of policy $\pi$, which can be (a function of) reward in RL. 
%Its constraint condition $\phi(\cdot)$ is determined by the following penalty function~\cite{runarsson2000stochastic}:
%\begin{equation}
%\phi(I) = \sum_i \max(0,J^{\pi}_{c_i}-\epsilon_i)^2.
%\end{equation}
Besides the principle of ranking the feasible solutions higher than infeasible solutions and ranking the infeasible solutions with higher penalty values lower, the core idea of stochastic ranking is randomly providing infeasible solutions with a chance to be ranked according to their fitness values only.

\section{evolutionary constrained Reinforcement Learning}

\begin{figure}[!h]
    \centering
    \includegraphics[width=1\linewidth]{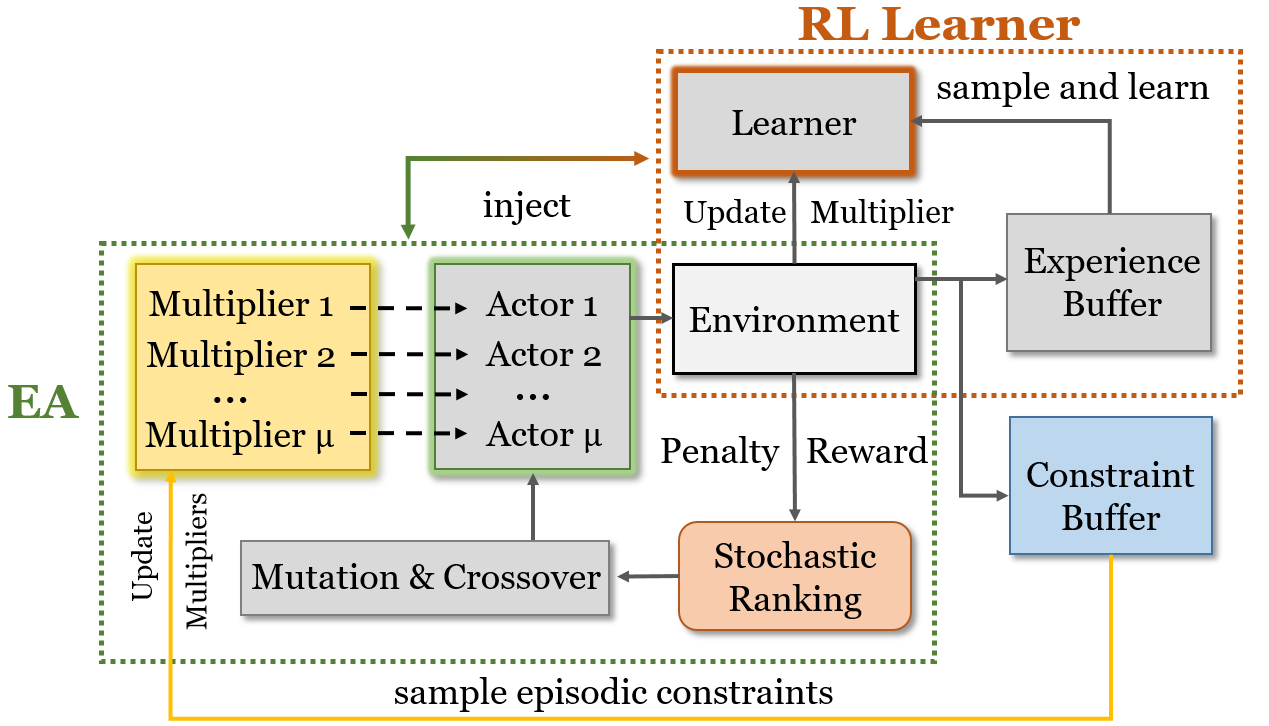}
    \caption{Architecture of ECRL. EA (green border) collects diverse experiences by actors after stochastic ranking. RL learner (red border) utilises the experiences to update the policy and injects the gradient into EA periodically. }
    \label{fig:ECRL}
\end{figure}

In this work, we propose evolutionary constrained reinforcement learning (ECRL), which incorporates, RCPO~\cite{tessler2018reward}, a constraint buffer of Lagrange relaxation coefficients~\cite{altman1999constrained}, and stochastic ranking~\cite{runarsson2000stochastic} into ERL~\cite{khadka2018evolution} to address constrained optimisation problems. Fig. \ref{fig:ECRL} illustrates the architecture of ECRL. 
%\subsection{Architecture of ECRL}
Algorithm \ref{alg:srerl} details the ECRL implemented with soft actor-critic (SAC)~\cite{haarnoja2018soft}. SAC is selected as the learning agent since it has been shown to achieve promising performance in related problems~\cite{haarnoja2018soft}. 

%\subsection{RCPO}

\begin{algorithm}[!ht]
\caption{ECRL using SAC. $Evaluate(\pi)$ evaluates a policy $\pi$ by interacting with a given environment.}
\label{alg:srerl}
%\color{blue}
\begin{algorithmic}[1] %[1] enables line numbers
\Require generation number $N$, population size $\mu$, learning rate of Lagrange multipliers $\eta$, mutation probability $p_m$, elite number $e$, synchronisation period $\omega$, SAC's temperature parameter $\alpha$%, constraint threshold $\epsilon$
\Ensure $\pi_\theta$
\State Initialise an experience replay buffer $\mathcal{B}_R$
\State Initialise a constraint replay buffer $\mathcal{B}_\sC$
\State Initialise a generational constraint buffer $\mathcal{B}^g_\sC$
%\State Initialise a constraint replay buffer $\mathcal{B}_C$
\State Initialise an RL agent $\pi_{\theta}$, its corresponding Lagrange multiplier $\lambda_{\pi_{\theta}}$, and two critics $\hat{Q}_{\psi_1}$ and $\hat{Q}_{\psi_2}$ % with related target networks
\State Initialise a population of $\mu$ individuals (i.e., actors) $\pi_1,\dots,\pi_\mu$ and their corresponding Lagrange multipliers $\lambda_1,\dots, \lambda_\mu$
\For{ $n=1$ to $N$}
    \For{$k=1$ to $\mu$}
     \State $J_\sR^{\pi_k},J_\sC^{\pi_k},\tau^{\pi_k},\lambda_k = Evaluate(\pi_k)$ %\Comment{$f$ is the fitness function. $\phi$ is the penalty function. $\mathcal{T}$ is the trajectory}
     \State Store $\tau^{\pi_k}$ and $\lambda_k$ into $\mathcal{B}_R$
     \State Store $J_\sC^{\pi_k}$ into $\mathcal{B}_\sC$ and $\mathcal{B}^g_\sC$
    \EndFor
    \State Sort $\pi_1,\dots,\pi_\mu$ with \emph{stochastic ranking} (Algorithm \ref{alg:sr}) according to
    $J_\sR^{\pi_1},\dots,J_\sR^{\pi_\mu}$ and $J_\sC^{\pi_1},\dots,J_\sC^{\pi_\mu}$
    % \State Select the first $e$ individual from the population
    \State Apply crossover operator and mutation operator with probability $p_m$ to the last $\mu-e$ individuals
    % \State  Apply mutation operator to the last $\mu-e$ individuals 
    % \State \warn{Apply genetic operators including mutation and crossover to the population $\pi_1,\dots,\pi_\mu$} 
    \State $J_\sR^{\pi_\theta},J_\sC^{\pi_\theta},\tau^{\pi_\theta},\lambda_{\pi_\theta} = Evaluate(\pi_\theta)$ %\Comment{$f$ is the fitness function. $\phi$ is the penalty function. $\mathcal{T}$ is the trajectory}
     \State Store $\tau^{\pi_\theta}$ and $\lambda_\pi$ into $\mathcal{B}_R$ 
     \State Store $J_\sC^{\pi_\theta}$ into $\mathcal{B}_\sC$ and $\mathcal{B}^g_\sC$
    \State Randomly sample a minibatch $B_R$ of transitions $\mathcal{T}= \langle s,a,s',r,c,\lambda\rangle$ from $\mathcal{B}_R$
    %\\\quad \quad \quad  $<s_t^i,a_t^i,r_t^i,c_t^i,\lambda^i,s_{t+1}^i> $ from $\mathcal{B}_R$
    \State Compute $y =r-\lambda c+\gamma (\min\limits_{j=1,2}\hat{Q}_{\psi_j}(s',\tilde{a}')-\alpha \log{\pi_\theta(\tilde{a}'|s')})$, where $\tilde{a}'\sim \pi_{\theta}(\cdot|s') $
    \State Update critic with 
    \NoNumber{$\nabla_{\psi_j}\frac{1}{|B_R|}\sum\limits_{\mathcal{T} \in B_R}^{}(y-\hat{Q}_{\psi_j}(s,a))^2$ for $j=1,2$}
    \State Update actor with 
    \NoNumber{$\nabla_{\theta}\frac{1}{|B_R|}\sum\limits_{\mathcal{T} \in B_R}(\min\limits_{j=1,2}\hat{Q}_{\psi_j}(s,\tilde{a}_{\theta})-\alpha \log\pi_\theta(\tilde{a}_{\theta}|s))$, $\tilde{a}_{\theta}$ is sampled from $\pi_\theta(\cdot|s)$ via reparametrisation trick}
    \State Apply soft update on target networks
    % \State Randomly sample a batch of $512$ transitions $<s_t^i,a_t^i,r_t^i,c_t^i,\lambda^i,s_{t+1}^i>$ from $\mathcal{B}_R$
    % %\\\quad \quad \quad  $<s_t^i,a_t^i,r_t^i,c_t^i,\lambda^i,s_{t+1}^i> $ from $\mathcal{B}_R$
    % \State Compute $y_i =r_t^i-\lambda^i c_t^i+\gamma (\min\limits_{j=1,2}\hat{Q}_{\psi_j}(s_{t+1}^i,\tilde{a}_{t+1})-\alpha \log{\phi(\hat{a}|s_{t+1}^i)})$, where $\tilde{a}_{t+1}\sim \pi_{\theta}(\cdot|s_{t+1}^i) $
    % \State Update critic with 
    % \NoNumber{$\nabla_{\psi_j}\frac{1}{|B_R|}\sum_i(y_i-\hat{Q}_{\psi_j}(s_t^i,a_t^i))^2$ for $j=1,2$}
    % \State Update actor with 
    % \NoNumber{$\nabla_{\theta}\frac{1}{|B_R|}\sum\nabla_{\theta} (\min\limits_{j=1,2}\hat{Q}_{\psi_j}(s_t^i,\tilde{a}_{t})-\alpha \log\pi(\tilde{a}_{t}|s_t^i))$, $\tilde{a}_{t}$ is sampled form $\pi_\theta(\cdot|s_t^i)$ via reparametrisation trick}
    \State Update Lagrange multiplier
    using Eq. \ref{eq:Ulambdapi}
    %with\Comment{\textit{Adopting Eq. \ref{eq:Ulambdapi}}}
    %\NoNumber{$\lambda_{\pi_\theta} = \max (\lambda_{\pi_\theta}+\sum\limits_{J_\sC^{\pi}\in \mathcal{B}^g_\sC}\eta(J_\sC^{\pi}-\epsilon),0)$} 
    \If{$n$ mod $\omega =0$}
    \State Randomly sample a batch $B_\sC$ from $\mathcal{B}_\sC$
    % \State $\hat{\lambda} = \frac{1}{|B|}\sum_i\lambda_i+\eta(J_{C}-\alpha)$
        \State Replace the lowest ranked individual $\pi_\mu$ with $\pi_\theta$ 
        \State Update its corresponding multiplier with $B_\sC$ \NoNumber{using Eq. \ref{eq:UlambdaE}}
        %\Comment{ Eq. \ref{eq:UlambdaE}}
       %\NoNumber{ $\lambda_\mu = \max(\lambda_\mu+\eta(\frac{1}{|B_C|}\sum\limits^{B_C}_{J^{B_C}_{C}}(J^{B_c}_{C}-\alpha)),0)$}
    \EndIf
    \State Empty the generational constraint buffer $\mathcal{B}^g_\sC$
\EndFor
% \State \textbf{return} solution
\end{algorithmic}
\end{algorithm}

RCPO is applied to the RL learner to consider constraint violations with the help of Lagrange relaxation multipliers $\lambda$. First, the constrained problem is converted to an unconstrained one using Lagrange relaxation method~\cite{tessler2018reward}, formulated as
\begin{equation}
\min_{\lambda}\max_{\theta}[J_{\sR}^\pi-\lambda (J_\sC^{\pi} - \epsilon)],
\end{equation}
where $\lambda>0$ is the Lagrange multiplier.
Then, we reshape the reward function with the penalty pattern~\cite{ng1999policy} as:
\begin{equation}\label{eq:reshapeR}
    \mathcal{R}'(s,a,s',\lambda) = \mathcal{R}(s,a,s')-\lambda c(s,a,s'). 
\end{equation}
\def\noused{So the value function can be formulated as Eq. (\ref{eq:V}) if we ignore the entropy pattern here for simplification.
\begin{equation}
\begin{aligned}
V^{\pi}&=\mathbb{E}^{\pi}[\sum \limits_{t=0}^{\infty}\gamma^t\hat{r}(s_t,a_t,\lambda)|s_0=s] \\
&=\mathbb{E}^{\pi}[\sum \limits_{t=0}^{\infty}\gamma^t(r(s_t,a_t)-\lambda c(s_t,a_t))|s_0=s]\\
&=V^{\pi}_{R}-\lambda V^{\pi}_\sC.
    \label{eq:V}
\end{aligned}
\end{equation}}

Besides the learner's multiplier, a Lagrange relaxation multiplier is maintained for each actor in the population (highlighted in yellow in Fig. \ref{fig:ECRL}). In addition to the experience buffer, a \emph{constraint buffer} $\mathcal{B}_\sC$ is used to store historical episodic constraints that are obtained from previous evaluations, aiming at updating the multipliers stably and efficiently.

ECRL first randomly initialises a population of actors and the corresponding multipliers. At each generation, experiences are sampled through actors' interactions in the environment. The transitions including corresponding multipliers will be added into an experience buffer $\mathcal{B}_R$, and the multipliers are also stored separately in the constraint buffer $\mathcal{B}_\sC$. 
%The constraint buffer collects historical episodic constraints that are obtained from previous evaluations and is expected to provide more stable and efficient update for multipliers.
%to update the corresponding multipliers.  \rew{The constraint buffer collects historical episodic constraints that are obtained from previous evaluations and is expected to provide more stable and efficient update for multipliers.}
% to store historical episodic constraints that are obtained from previous evaluations.
Actors in the population are ranked by \emph{stochastic ranking} (Algorithm \ref{alg:sr}) according to their fitness (i.e., reward) and penalties. 
%$f(\cdot)$ in Algorithm~\ref{alg:sr} is set as the discounted cumulative reward in Algorithm~\ref{alg:srerl}.
Following \citet{runarsson2000stochastic}, the penalty for constraint violations in Algorithm~\ref{alg:sr} can be determined by 
\begin{equation}
\phi(\pi)=\max(0,J^{\pi}_\sC-\epsilon)^2.
%\phi(\pi)=\sum_i \max(0,J^{\pi}_{C_i}-\epsilon_i)^2.
\end{equation}
Notably, our approach can handle multiple constraints in a way similar to stochastic ranking, where the penalty function is the sum of the quadratic loss of all constraints, $\phi(\pi)=\sum_{i=1}^m max(0, J^{\pi}_{C_i}-\epsilon_i)^2$, considering $m$ constraints.
Actors in the population are ranked by \emph{stochastic ranking} (Algorithm \ref{alg:sr}) as follows according to their fitness and penalties.
The mutation and crossover
%and elite survival 
operators of ERL are reserved. 
%However, before performing any genetic operators, actors in the population are ranked by \emph{stochastic ranking} (Algorithm \ref{alg:sr}) as follows according to their fitness and penalties. 
%Assuming a solution is a policy in RL, 
%the following penalty function~\cite{runarsson2000stochastic}: % using previously defined notations:
%\begin{equation}
%\phi(\pi) = \sum_i \max(0,J^{\pi}_{c_i}-\epsilon_i)^2.
%\end{equation}
Then, current episodic constraint values sampled by the ranked population are used to update the learner's multiplier, following
\begin{equation}\label{eq:Ulambdapi}
%\color{blue}
    \lambda_{\pi} = \max (\lambda_{\pi}+\eta(J_\sC^{\pi}-\epsilon),0). 
    % \hat{\lambda} = \max (\mathrm{E}(\lambda_(pop))+\eta(\mathrm{E}(\phi)),0)
\end{equation}
When injecting the learner's gradient information into the EA's population, the lowest ranked actor's Lagrange multiplier, $\lambda_\mu$, is updated with a batch ${B_\sC}$ sampled from the constraint buffer $\mathcal{B}_\sC$, using
\begin{equation}\label{eq:UlambdaE}
%\color{blue}
\lambda_\mu = \max (\lambda_\mu +\eta\left(\frac{1}{|B_\sC|}\sum\limits_{J^{\pi}_\sC \in {B_\sC}}(J^{\pi}_{\sC}-\epsilon)\right),0), 
\end{equation}
where $\eta$ is the learning rate of multipliers.

ECRL considers constraints with Lagrange relaxation multipliers and applies stochastic ranking to balance rewards and penalties in EA. The multipliers provide diverse experiences considering constraint information to restrict RL learner and are updated stably with the help of the constraint buffer. It is believed that the proposed method tackles the conflicting behaviour of EA and RL and further handles CRL problems.
%and $B_C$ is a mini batch sampled from constraint buffer $\mathcal{B}_C$.

% It is worth to mention that ECRL requires no priori knowledge.

% \todo[inline]{ I am not sure if including the constraint buffer in this paper. This is true that it is sort of minor . But I think this can be applied to almost augmenting based approaches. Can we split it out for anthor paper?}

\begin{algorithm}[t]
\caption{\label{alg:sr}Stochastic ranking \cite{runarsson2000stochastic} in ECRL. $J_\sR^{\pi}$ and $\phi(\pi)$ denote the reward and penalty of $\pi$ considering one or multiple constraints, respectively. $P_f\in (0,1)$ is the tolerate probability.}
\begin{algorithmic}[1] 
\Require a population of $\mu$ individuals $\pi_1,\cdots,\pi_\mu$
\Ensure $\pi_{I_1},\cdots,\pi_{I_\mu}$ after sorting
\State $I_j = j, \forall j \in \{1, \dots,\mu\}$
\For{$i=1$ to $\mu$}
    \For{$j=1$ to $\mu-1$}
        \State Sample $\zeta$ uniformly at random in $(0,1)$ \label{line:random}
        %J_\sR^{\pi_\theta},J_{C}^{\pi_\theta}
        \If{$(\phi(\pi_j)=\phi(\pi_{j+1})=0)$ or $(\zeta < P_f)$} 
           \If{$J_\sR^{\pi_j}<J_\sR^{\pi_{j+1}}$}
            \State swap $I_j$ and $I_{j+1}$
        \EndIf \label{line:endrandom}
        \Else
                \If{$\phi(\pi_j)>\phi(\pi_{j+1})$}
            \State swap $I_j$ and $I_{j+1}$
                    \EndIf
        \EndIf
    \EndFor
\EndFor
\end{algorithmic}
\end{algorithm}

\def\old{
\begin{algorithm}[!ht]
\caption{\label{alg:sr}Stochastic ranking~\cite{runarsson2000stochastic} in ECRL. $P_f\in (0,1)$ is the tolerate probability. $(\cdot)$ and $\phi(\cdot)$ denote the fitness function and penalty function considering one or multiple constraints, respectively.}
\begin{algorithmic}[1] 
\Require a population of $\mu$ individuals $I_1,\cdots,I_\mu$
\Ensure $I_1,\cdots,I_\mu$ after sorting
\For{$i=1$ to $\mu$}
    \For{$j=1$ to $\mu-1$}
        \State Sample $\zeta$ uniformly at random in $(0,1)$ \label{line:random}
        %J_\sR^{\pi_\theta},J_{C}^{\pi_\theta}
        \If{$(\phi(I_j)=\phi(I_{j+1})=0)$ or $(\zeta < P_f)$} 
           \If{$f(I_j)<f(I_{j+1})$}
            \State swap $I_j$ and $I_{j+1}$
        \EndIf \label{line:endrandom}
        \Else
                \If{$\phi(I_j)>\phi(I_{j+1})$}
            \State swap $I_j$ and $I_{j+1}$
                    \EndIf
        \EndIf
    \EndFor
\EndFor
\end{algorithmic}
\end{algorithm}

}

\section{Experiments}\label{sec:exp}

We conduct several sets of experiments over continuous robotic control benchmarks with the torque constraint to (i) demonstrate the limitation of directly applying CRL approaches to ERL to address constrained problems and provide experimental evidence to support our motivation for ECRL,
%, including the tradeoff between reward and constraint violations when directly applying the classic reward shape validation to ERL, 
% demonstrate the tradeoff between reward and constraint violations when directly applying the classic reward shape validation to ERL,
(ii) validate the performance of ECRL by comparing it to several state-of-the-art algorithms, and (iii) show the contributions of stochastic ranking and constraint buffer of Lagrange multipliers via an ablation study.

\subsection{Continuous Control Tasks in Mujoco}
\begin{figure}[t]
    \centering
    \includegraphics[width=0.35\columnwidth]{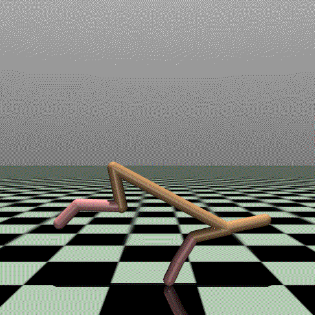}\hspace{1.5em}
	\includegraphics[width=0.35\columnwidth]{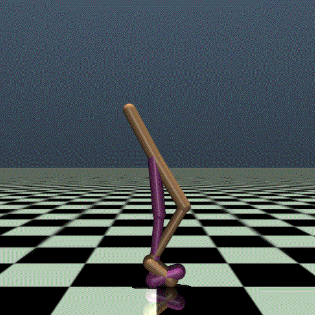}
	\caption{\label{fig:menv}Screenshots of \emph{HalfCheetah} (left) and \emph{Walker2d} (right)~\cite{todorov2012mujoco}.
 %\footref{fn:mujoco}.
 }
\end{figure}

OpenAI Gym environment~\cite{OpenAIGym} integrated with the MuJoCo simulator is used, considering five robot control tasks that are widely used in the literature~\cite{todorov2012mujoco}, namely \emph{Ant}, \emph{HalfCheetah}, \emph{Walker2d}, \emph{Hopper} and \emph{Swimmer}. As illustrative examples, Fig. \ref{fig:menv} shows the environments of \emph{HalfCheetah} and \emph{Walker2d}.
%\footnote{\label{fn:mujoco}\url{https://www.gymlibrary.ml/environments/mujoco/}}. 

All robots in those tasks consist of joints, legs and torso and are controlled by applied torques. All of the five tasks aim to maximise the gained reward, in terms of walking performance, under a torque constraint.
% The five tasks are described as follows.
% \begin{itemize}
%     \item \emph{Ant}: The three-dimensional Ant robot with four legs is rewarded by moving forward.
%     \item \emph{HalfCheetah}: The HalfCheetah is a two-dimensional robot consisting of nine links and eight joints and rewarded by running forward.
%     \item \emph{Walker2d}: The walker is a two-dimensional two-legged figure that consists of four main body parts and is rewarded by running forward.
%     \item \emph{Hopper}: The hopper is a two-dimensional one-legged figure that consists of four main body parts and is rewarded by the forward movement.
%     \item \emph{Swimmer}:The swimmer consists of three links and rotors. The goal of the robot is to move as fast as possible.
% \end{itemize}
Tab. \ref{tab:InfoM} summarises the dimension of the observation space and action space in each task. More details can be found in the work of \citet{todorov2012mujoco}. 

The constraint threshold $\epsilon$ is set as $0.4$. Specifically, the average torque applied to each motor is defined as a constraint $C$, and the per-step average torque is $c(s,a,s')$, following
\begin{equation}\label{eq:crobot}
C = \frac{1}{|\tau|}\sum_{t=0}^{|\tau|-1}c(s_t,a_t,s_{t+1}) \leq \epsilon,
\end{equation}
where $\tau$ denotes a sampled trajectory $(s_0,a_0,s_1,a_1,\dots)$.

%where $\tau$ is a sampled trajectory.
%\todo[inline]{$\mathcal{T}$ is not defined}
% % \subsubsection{Humanoid}
% The 3D bipedal robot with 17 controllable joints is designed to walk forward without falling over.
\def\toolong{\subsubsection{Ant }
The three-dimensional Ant robot with four legs is rewarded by moving in forward.
\subsubsection{Hopper }
The Hopper is a two-dimensional one-legged figure that consists of four main body parts and is rewarded by the forward movement.
\subsubsection{HalfCheetah } The HalfCheetah is a 2-dimensional robot consisting of 9 links and 8 joints connecting them and rewarded by running forward.
\subsubsection{Walker2d } The Walker is a two-dimensional two-legged figure that consists of four main body parts and is rewarded by running forward.
\subsubsection{Swimmer } The Swimmer consists of several links and rotors. The goal of the robot is to move as fast as possible.}

\begin{table}[htbp]
    \caption{Summary of considered continuous control tasks. Value of an action's each dimension ranges in $[-1,1]$.}
    \centering
    \begin{tabular}{c|c|c}%|c}
    \toprule
         Task & Observation dimension & Action dimension \\%& $\alpha$\\
         \midrule
        %  Humanoid& 376 &17 & -\\
         \emph{Ant} & 27 & 8 \\% & 0.4 \\
         \emph{HalfCheetah} & 17& 6\\ % & 0.4 \\
         \emph{Walker2d} & 17 & 6\\ % & 0.4 \\
         \emph{Hopper} & 11& 3 \\% & 0.4 \\
         \emph{Swimmer} & 8 & 2\\ % & 0.4 \\
         \bottomrule
    \end{tabular}

    \label{tab:InfoM}
\end{table}
\begin{figure*}[!h]
    \centering
    \includegraphics[width=0.9\linewidth]{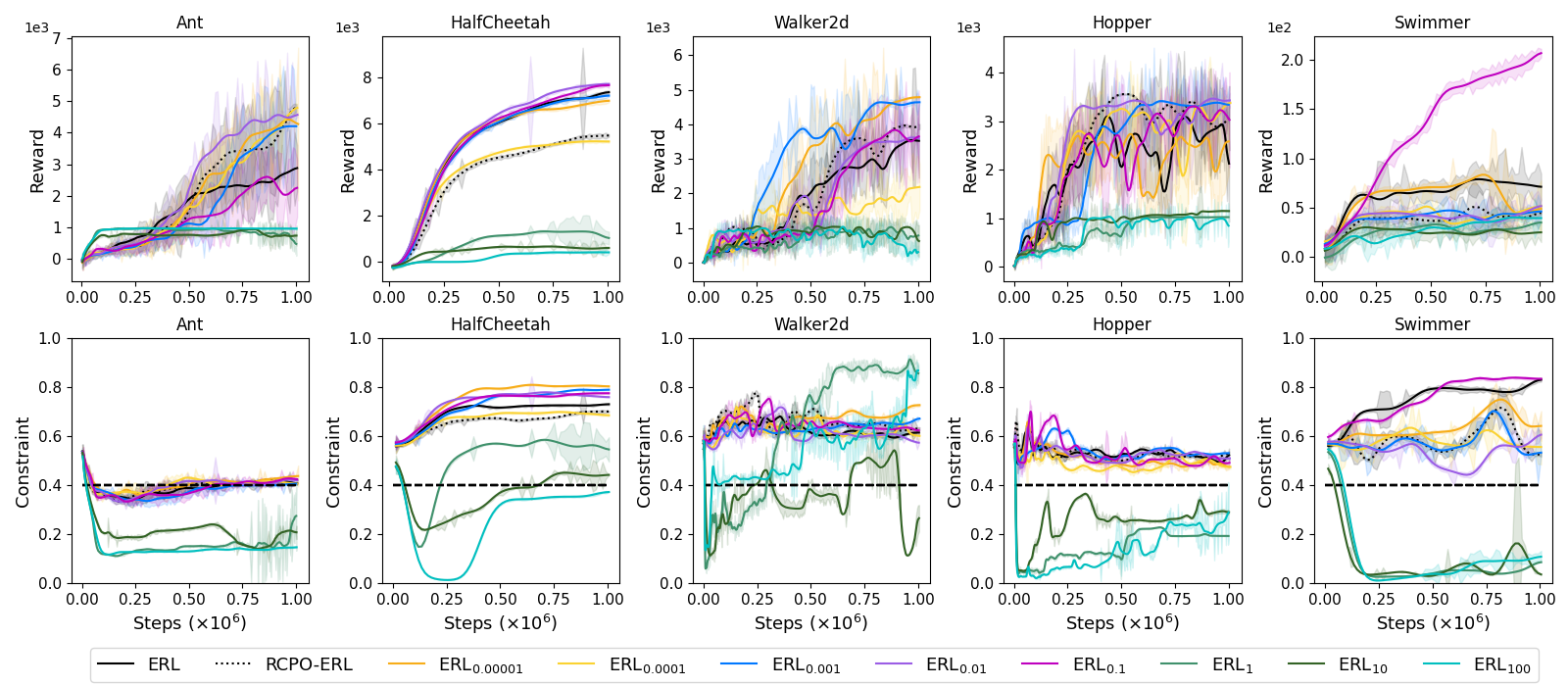}
    \caption{Learning curves of reward (top) and constraint violation (bottom) of ERL, RCPO-ERL and ERL with reward shape using different penalty coefficient values on five Mujoco control tasks.
    %Two rows of figures show rewards and constraints from top to bottom, respectively.
    The black dashed line refers to the torque constraint $\epsilon=0.4$. Curves below the black dashed line imply constraint satisfaction.}
    \label{fig:reshape}
\end{figure*}

\subsection{Compared Algorithms}
ECRL is compared to several state-of-the-art algorithms including ERL~\cite{khadka2018evolution}, RCPO~\cite{tessler2018reward}, and IPO~\cite{liu2020ipo}. In addition, we implement two variants of ERL with RCPO~\cite{tessler2018reward} and a reward shape method~\cite{ng1999policy} using different coefficient values for constrained optimisation, as compared algorithms.

RCPO implemented with SAC is denoted as ``RCPO'' and IPO implemented with PPO is denoted as ``IPO'' in the experiment results, respectively.
%SAC integrated with RCPO is denotes as ``RCPO''.
Adopting RCPO to ERL is straightforward and not detailed here. The resulted algorithm is referred to as ``RCPO-ERL''. When adopting the reward shape method to ERL, it is hard to determine the best value for penalty coefficient $\lambda$. In addition to $0.00001$, $1$ and $100$ tested in previous work~\cite{tessler2018reward}, we also test $\lambda=0.0001$, $0.001$, $0.01$, $0.1$, $1$ and $10$ to demonstrate the impact brought by the choice of penalty coefficient $\lambda$. Instances of ERL with reward shape using different values of $\lambda$ are referred to as ``$\text{ERL}_{*}$'', where $*$ indicates the value of penalty coefficient $\lambda$. ``$\text{ERL}$'' refers to the original ERL without reward shaping or RCPO.
%\subsubsection{RCPO}
%\subsubsection{ERL with Shaped Reward}

%We adapt the classic reward shape method~\cite{ng1999policy} to ERL for constrained optimisation. 
%The reward shape method~\cite{ng1999policy} is a classic and easy-to-implement method for constrained optimisation. It shapes the reward function with a weighted penalty term~\cite{ng1999policy}, as shown in Eq. \ref{eq:reshapeR}. 

%\subsubsection{ERL with RCPO}
%We also adapt RCPO~\cite{tessler2018reward}, a state-of-the-art Lagrange relaxation based method with multi-timescale update to ERL.

\subsection{Experiment Setting}
The network structure used by all the compared algorithms is a fully connected neural network with a $\langle256,256\rangle$ linear layer. The temperature parameter of SAC $\alpha$ is set as 0.1. The learning rate of actor and critic are set as 1e-4 and 3e-4, respectively. The size of the experience replay buffer and experience batch are set as 1e6 and 512, respectively. 
% The discounted factor $\gamma$ is 0.99. 

ECRL (Algorithm \ref{alg:srerl}) is implemented with Tianshou platform~\cite{tianshou} and integrated with a parallel version of original ERL\footnote{\label{a} Source Codes of this work are available at \url{https://github.com/HcPlu/Evolutionary-Constrained-Reinforcement-Learning}}. RCPO and IPO are implemented by us since the relevant codes are not distributed.
The learner's multiplier is initialised as $0.001$ and its learning rate $\eta$ is 1e-5 for RCPO, RCPO-ERL and ECRL. The size of the constraint buffer and batch are set as 100 and 32, respectively. The synchronisation period $\omega$ for injecting gradient information is set as $1$. 
% The population size $\mu$ and elite number $e$ are 10 and 2, respectively. 
The mutation probability $p_m$ is 0.9. The multipliers $\lambda_1, \dots,\lambda_\mu$ are sampled uniformly at random from $(0,1)$ for initialisation. Logarithmic barrier function of IPO' parameter $t$ is set as 50. Other hyperparemters of IPO can refer to~\cite{tianshou}. The parameter values are either set following previous studies~\cite{tianshou,khadka2018evolution} or arbitrarily chosen. All algorithms are trained for 1e6 timesteps on the same machine with 4 v100 GPU cores. Five independent runs are applied each time for test during training.

\def\toolong{
\begin{table}[htbp]
    \centering
    \begin{tabular}{c|c}
    \toprule
        Parameters & value  \\
        \midrule
       $\#$ population  & 10\\
       $\#$ rollout & 5 \\
       Timesteps & 1e6\\
       $\gamma$ & 0.99\\
       Learning rate of actor & 1e-4\\
       Learning rate of critic & 3e-4\\
       Experience buffer size & 1e6 \\
       Experience batch size & 512 \\
       Constraint buffer size & 100\\
       Constraint batch size & 32 \\
       \bottomrule
    \end{tabular}
    \caption{Some hyperparameters for ECRL.}
    \label{tab:pms}
\end{table}}

\subsection{Challenges of Setting Good Penalty Coefficients}
%\rew{
Experiment results of ERL, ERL with RCPO, and ERL with the reward shape method using different penalty coefficient values are presented in Fig. \ref{fig:reshape}
%to the five robot control tasks 
to support our motivation of introducing stochastic ranking and constraint buffer.
% We detail the motivation of designing ECRL with the experiment results of applying the original ERL and ERL with reward shape method to the five robot control tasks.
%Fig. \ref{fig:reshape} provides the training curves of the original ERL, \warn{RCPO} and the ERL with reward shape using different penalty coefficient values.}
%\todo[inline]{Add RCPO-ERL to Fig. \ref{fig:reshape}}
%, referred to as $\text{ERL}_{*}$, where $*$ indicates the value of penalty coefficient $\lambda$. 

\begin{figure*}[ht]
    \centering
    \includegraphics[width=0.9\linewidth]{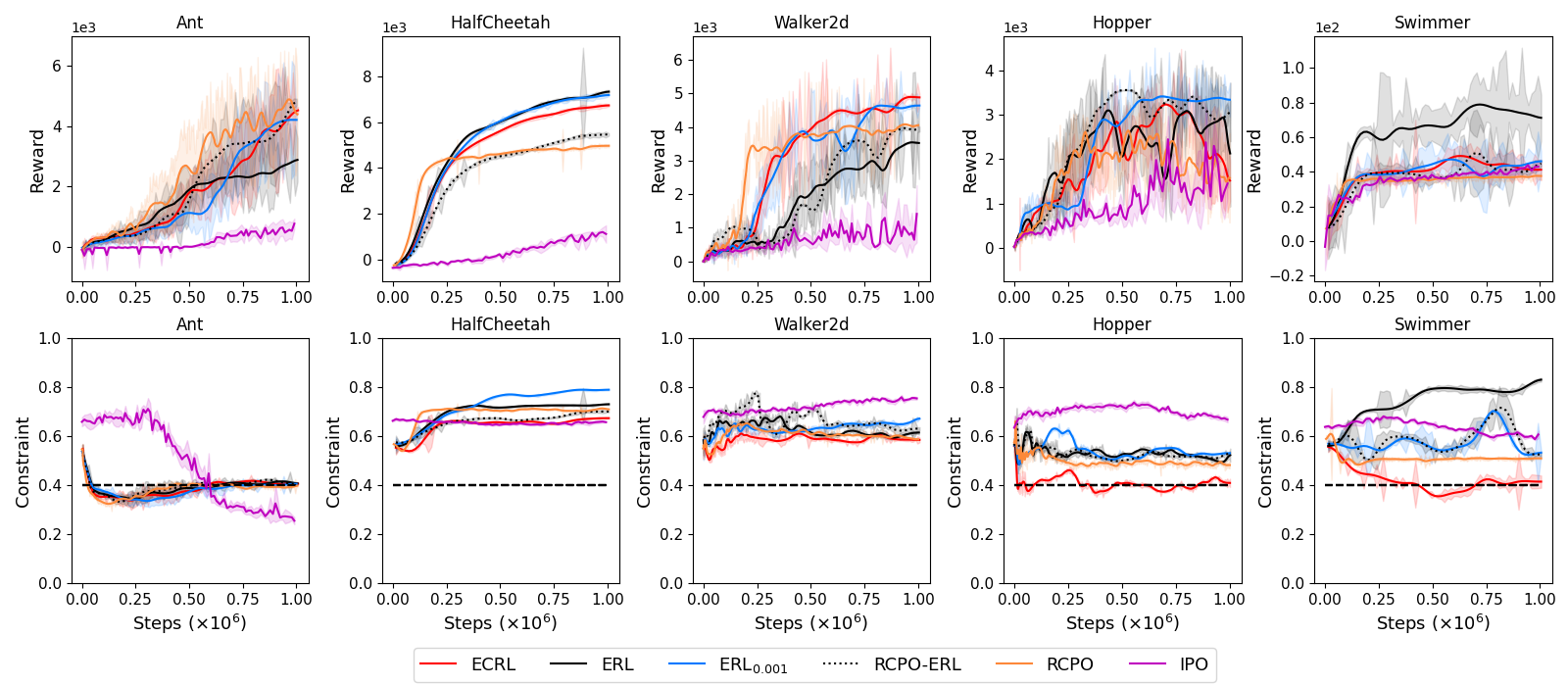}
    \caption{Learning curves of reward (top) and constraint violation (bottom) of ECRL, RCPO, IPO, RCPO-ERL, ERL with reward shaping and ERL. The black dashed line refers to the torque constraint $\epsilon=0.4$.}
    \label{fig:com}
\end{figure*}

\subsubsection{Dilemma between Reward and Constraint Violation}
As shown in Fig. \ref{fig:reshape}, shaping the reward function properly is not trivial. There is no universally good penalty coefficient value for all tasks. For example, ERL agents with fixed small multipliers like 0.00001 and 0.0001 can score about 4000 in \emph{Ant} and 7500 in \emph{HalfCheetah}. The ERL agent with $\lambda=0.1$ can score 200 in \emph{Swimmer}. Prior knowledge or fine tuning is needed to set a suitable $\lambda$ value. As observed from experiment results, using a smaller penalty coefficient ($\lambda\leq 0.1$) often leads to higher reward and constraint violation, while using a larger penalty coefficient ($\lambda=1$, $10$ and $100$) usually leads to smaller constraint violation but worse performance in terms of reward. 
However, one hardly concludes that the policy will necessarily satisfy the constraint by varying the penalty coefficient. We can only state that with a high probability, this trend exists due to the variance of the optimal policy.
The dilemma between reward and constraint violation is observed when directly applying the reward shape method to ERL.

\subsubsection{Conflicting Behaviour of EA and RL} Fig. \ref{fig:reshape} indicates that when using larger $\lambda$ (cf. $\text{ERL}_{1}$, $\text{ERL}_{10}$ and $\text{ERL}_{100}$), the constraint violation may increase along with training, especially in \emph{HalfCheetah}, \emph{Walker} and \emph{Hopper}. Taking \emph{HalfCheetah} as an example, at the early stage, these learning curves have a fast decline on constraint violation, then between timestpes 1e5 and 5e5, their constraint violations increase along with the augmentation of reward. This phenomenon is explained by the conflicting behaviour of EA and RL. When shaping the reward function with constraint violation using Eq. (\ref{eq:reshapeR}), the trained agent is expected to be guided to learn the policy that maximises the reward and minimises the penalty due to constraint violation until satisfying constraints. However, it is notable that an agent learns from experiences sampled by actors in the population of ERLs, which do not consider constraint. Although the gradient information is injected into the population periodically, it still fails due to the high discard rate of elite and fitness-guided selection~\cite{khadka2018evolution} that relies on reward only. More specifically, EA and RL sample from two different distributions with different concerns, one over reward only and the other over both reward and constraint. When the majority of the experience buffer is replaced with reward-only experiences, the gradient will lead to the direction of optimising reward only, which may violate constraints.

The dilemma between the reward and constraint violation as well as the presence of the conflicting behaviour of EA and RL suggest that simply applying CRL techniques to ERL is not enough. This motivates us to propose ECRL which leverages stochastic ranking and Lagrange relaxation.

\begin{figure*}[ht]
    \centering
    \includegraphics[width=0.9\linewidth]{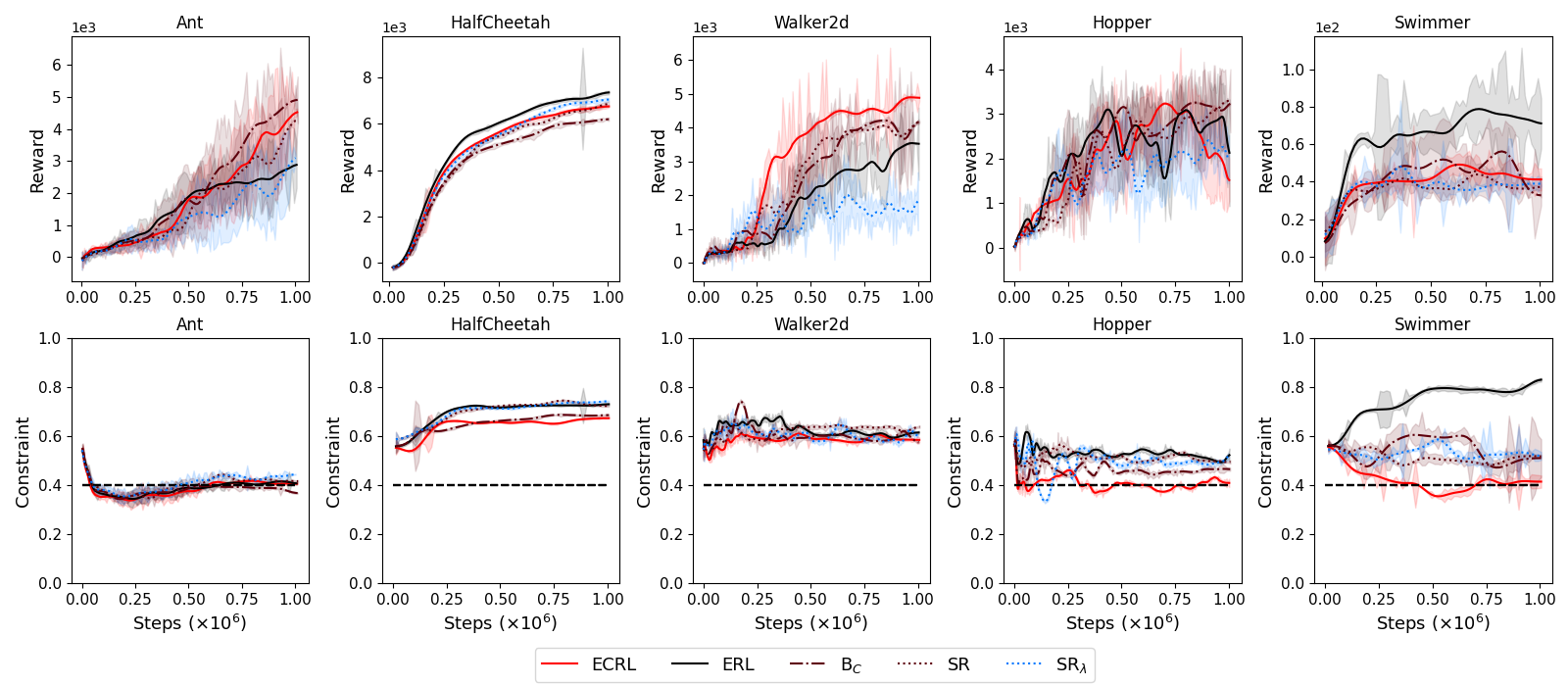}
    \caption{Results of ablation study. 
    %Stochastic ranking and constraint buffer are split out from ECRL for ablation. Two rows of figures show rewards and constraints from top to bottom, respectively.
    The black dashed line refers to the torque constraint $\epsilon=0.4$.}
    \label{fig:ablationECRL}
\end{figure*}

\subsection{Comparing ECRL with State-of-the-arts}
% \todo[inline]{add ECRL with 0.001, 1, 10}
% \begin{table*}[]
%     \centering
%     \begin{tabular}{c|cc|cc|cc|cc|cc}
%     \toprule
%          \multicolumn{1}{c|}{\multirow{2}{*}{Agent}}&\multicolumn{1}{c|}{\emph{Ant}}&\multicolumn{1}{c|}{\emph{Ant}}&\multicolumn{1}{c|}{\emph{Ant}}&\multicolumn{1}{c|}{\emph{Ant}} &\multicolumn{1}{c|}{\emph{Ant}} \\
%          \midrule
%          \bottomrule
%     \end{tabular}
%     \caption{Caption}
%     \label{tab:com}
% \end{table*}

To demonstrate the outstanding performance of our approach, we compare ECRL with state-of-the-art algorithms for CMDP including RCPO~\cite{tessler2018reward}, IPO~\cite{liu2020ipo}, ERL~\cite{khadka2018evolution}, RCPO-ERL and ERL$_{0.001}$ (the ERL with the reward shaping method using the best constraint coefficient value tested above). Fig. \ref{fig:com} presents the learning curves.
%, and the aforementioned variants of ERL for CMDP. 

According to Fig. \ref{fig:com}, it is clear that ECRL outperforms the compared algorithms. In the task \emph{Ant}, all the agents learned the policies that satisfy the torque constraint, but a clear difference is observed in terms of reward between them. ECRL, RCPO-ERL, RCPO and $\text{ERL}_{0.001}$ perform similarly and better than ERL with 2000 in terms of reward. IPO gets the lowest reward with constraint satisfaction. Although no agent satisfies the constraint in \emph{HalfCheetah} or \emph{Walker2d}, our ECRL agent still obtains the lowest constraint violation and promising performance in terms of reward. In \emph{Hopper} and \emph{Swimmer}, only our ECRL agent restricts policy with constraint satisfaction. 

%As excepted, ERL performs the worst on constraint among these agents since it does not consider subjecting to the limitation of torques. So it can get the highest score in \emph{HalfCheetah} by large movement and speed. 

% The major cause of the conflicting behaviour of EA and RL is the failure of incorporating RCPO and reward shape method into ERL.
ECRL coordinates EA and RL by the incorporation of stochastic ranking and Lagrange relaxation method to address CRL problems. Stochastic ranking improves the selection behaviour of actors in the population. Constrained policies can get balanced according to their episodic reward and constraint value so that the percentage of experiences considering constraint information in the experience buffer can increase. Naturally, the learner's gradient information samples and learns from these experiences for further improvement. Moreover, the constraint buffer provides historical constraint values for the stable update of the population's multipliers which is coordinated with the learner's multiplier. Benefited from this, ECRL tackles the innate conflict of EA and RL, as well as the dilemma between reward and constraint violations. 
ECRL achieves superior performance on the five continuous robotic control tasks than the compared state-of-the-art algorithms.
%\rew{Promising better performance  over the five \warn{continuous} robotic control problems are achieved by ECRL, comparing with state-of-the-arts.}

\subsection{Ablation Study}
An ablation study is also performed to validate the contributions of stochastic ranking and the constraint buffer. Fig. \ref{fig:ablationECRL} presents the experiment results. ``SR'' refers to the variant of the ERL using stochastic ranking only, thus no Lagrange multiplier or constraint buffer is used.
``SR$_\lambda$'' refers to the variant using stochastic ranking with penalty coefficient $\lambda=0.001$ without maintaining a constraint buffer.
``B$_\sC$'' refers to the variant using Lagrange multipliers and  constraint buffer, but no stochastic ranking. 
The values of all related common hyperparameters stay the same for a fair comparison.

According to Fig. \ref{fig:ablationECRL}, the constraint violation of our ECRL agent is the lowest in four out of five tasks.
%except \emph{Ant}. 
Particularly, in \emph{Walker2d}, ECRL outperforms the other algorithms for nearly 5000 in terms of reward. Although ECRL does not show significantly better convergence in terms of reward in \emph{Hopper} and \emph{Swimmer}, it restricts the policy with constraint satisfaction, while the others don't. All agents perform similarly in terms of constraint violation in \emph{Ant}.
% In \emph{HalfCheetah}, SR has a similar constraint value with ERL. 
% And SR even is worse than ERL in \emph{Ant} and \emph{Walker2d} on constraint. It is in \emph{Hopper} and \emph{Swimmer} that SR shows a significant difference with ERL. For CB agent, it gets good performance on reward, especially in \emph{Swimmer}, about 200 score is obtained which outperforms other agents. However, the price for such an high score is a large constraint value closing to 1, the maximal constraint value. 
% \subsubsection{Coordination of ECRL}
% Promising performance of ECRL is observed.
Comparing SR with B$_\sC$, neither behaves well over the tasks except \emph{Ant}, even both perform better than ERL. It can be blamed on the conflicting behaviour of EA and RL for the failure of using stochastic ranking or constraint buffer alone. 
%, even they perform better than ERL
%\warn{For stochastic, the selection of actor gets balanced but these is no direct gradient information containing constraint. Stochastic ranking is only responsible for choosing suitable actor for surviving. What stands in the way is that this selected actor will not guide the training procedure of learning directly. Even though experience sampled by balanced actors may contain guided direction for optimising constraint policy, the obstacle RL caused still exists. The learner learns from these experiences with constraint information but optimise the policy with maximising reward without considering constraint. Similar case happens to CB agent. The direction of optimisation is turned to not only maximise discounted cumulative reward but also restrict the behaviour of the agent with help of the Lagrange multiplier. However, selection of ERL only ranks actor according its gained reward. Thus, during training, reward-guide experiences will be the main component of experience buffer where a conflict raises.}

Stochastic ranking has an indirect impact on the learner. The constrained RL policy is guided by the experiences that are sampled by the ranked actors by stochastic ranking. Although those experiences may contain some useful constraint-related information, they contradict the learner's goal of maximising the reward and can hardly make a direct impact without a penalty coefficient on value function and action policy.

A similar situation happens to the B$_\sC$ agent. The behaviours of an RL agent are restricted by the Lagrange multiplier. However, the selection mechanism of ERL only ranks actors according to their rewards. Thus, reward-guided experiences become the main component of the experience buffer during training when a conflict raises again. Although the learner tries to optimise their policy under the reshaped value function, it is burdened by reward-guided experiences. 
The constraint buffer stores recent constraint values sampled by actors in the population and RL agent. The batch sampled from the constrained buffer is a collection of constraint values of the most recent searched policies, which indicates information on the decreasing direction of constraint value. With this information, it is possible to update the multipliers more efficiently and smoothly and guide the policy towards the constrained region.

 Stochastic ranking, constraint buffer and multipliers together form the bridge of communication and are indispensable according to our ablation study. Stochastic ranking and constraint buffer endow EA with the ability to sample diverse experiences considering both reward and constraint. RL learner restricts its behaviours with the help of Lagrange multipliers and experiences sampled by actors in EA. 

\section{Conclusion}
\label{sec:con}
In this paper, we propose an evolutionary constrained reinforcement learning (ECRL) algorithm to address constrained optimisation problems while balancing well the reward and constraint violation. ECRL applies stochastic ranking to balance the reward and constraint violation of actors in the population. The constraint buffer comprised of Lagrange multipliers further enhances ECRL. The tradeoff between reward and constraint violation and the conflict between EA and RL are studied in depth. Experiment results show that ECRL outperforms state-of-the-art algorithms on continuous robotic control benchmarks with the help of coordination between stochastic ranking and Lagrange multipliers that are stably updated by the constraint buffer. The contributions of ECRL's core ingredients are validated with an ablation study. This work further provides an understanding of the mechanism and interactions between EA and RL in addressing constrained problems. It can be a good start to reconsidering utilising traditional constraint handling techniques in EA for CRL problems. 
%ECRL achieves promising performance in solving robotic control benchmarks with defined constraints. Further analysis on the effectiveness of the proposed approach on more complex-constrained is needed. 
Future work will focus on applying ECRL to more real-world problems with constraints and validate its abilities and weaknesses.

\footnotesize{
\bibliographystyle{IEEEtranN} 
\bibliography{main}
}

\end{document}